\documentclass[runningheads]{llncs}
\usepackage{graphicx}
\usepackage{amsmath,amssymb} 
\usepackage{color}

\usepackage{times}
\usepackage{epsfig}
\usepackage{graphicx}
\usepackage{amsmath}
\usepackage{amssymb}
\usepackage{bm}
\usepackage{xcolor}
\usepackage{tabu}
\usepackage{color,array}
\usepackage{multirow,makecell}
\usepackage{subfig}

\begin{document}
\title{Maximum Margin Metric Learning Over Discriminative Nullspace for Person Re-identification} 

\titlerunning{Maximum Margin Metric Learning}

\author{T M Feroz Ali \inst{1} \and Subhasis Chaudhuri\inst{1} }
\authorrunning{Feroz Ali and S. Chaudhuri}

\institute{Indian Institute of Technology Bombay, Mumbai, India \and
\email{\{ferozalitm,sc\}@ee.iitb.ac.in}}
\maketitle             
\begin{abstract}
In this paper we propose a novel metric learning framework called Nullspace Kernel Maximum Margin Metric Learning (NK3ML) which efficiently addresses the small sample size (SSS) problem inherent in person re-identification and offers a significant performance gain over existing state-of-the-art methods. Taking advantage of the very high dimensionality of the feature space, the metric is learned using a maximum margin criterion (MMC) over a discriminative nullspace where all training sample points of a given class map onto a single point, minimizing the within class scatter. A kernel version of MMC is used to obtain a better between class separation. Extensive experiments on four challenging benchmark datasets for person re-identification demonstrate that the proposed algorithm outperforms all existing methods. We obtain 99.8\% rank-1 accuracy on the most widely accepted and challenging dataset VIPeR, compared to the previous state of the art being only 63.92\%.\footnote{\textit{Accepted to European Conference on Computer Vision (ECCV) 2018}}

\end{abstract}
\section{Introduction}
Person re-identification (re-ID) is the task of matching the image of pedestrians across spatially \color{black}non overlapping cameras, even if the pedestrian identities  are \textit{unseen} before. It is a very challenging task due to large variations in illumination, viewpoint, occlusion, background and pose changes. Supervised methods for re-ID generally include two stages: computing a robust feature descriptor and learning an efficient distance metric. Various feature descriptors like SDALF\cite{SDALF}, LOMO\cite{LOMO} and GOG\cite{GOG} have improved the efficiency to represent a person. But feature descriptors are unlikely to be completely invariant to large variations in the data collection process and hence the second stage for person re-identification focusing on metric learning is very important. They learn a discriminative metric space to minimize the intra-person distance while maximizing the inter-person distance. It has been shown that learning a good distance metric can drastically improve the matching accuracy in re-ID. Many efficient metric learning methods have been developed for re-ID in the last few years, for e.g., XQDA\cite{LOMO}, KISSME\cite{KISSME},  LFDA\cite{LFDA:CVPR}. However, most of these methods suffer from the small sample size (SSS) problem inherent in re-ID since the feature dimension is often very high.

Recent deep learning based methods address feature computation and metric learning jointly for an improved performance. However, their performance depends on the availability of manually labeled large training data, which is not possible in the context of re-ID. Hence we refrain from discussing deep learning based methods in this paper, and concentrate on the following problem: given a set of image features, can we design a good discriminant criterion for improved classification accuracy for cases when the number of training samples per class is very minimal and the testing identities are unseen during training. \color{black}Our application domain is person re-identification.

In this paper we propose a  novel metric learning framework called Nullspace Kernel Maximum Margin Metric Learning (NK3ML) which efficiently addresses the SSS problem and provide better performance compared to the state-of-the-art approaches for re-ID. The discriminative metric space is learned using a maximum margin criterion over a discriminative nullspace. In the learned metric space, the samples of distinct classes are separated with maximum margin while keeping the samples of same class collapsed to a single point (i.e., zero intra-class variance) to maximize the separability in terms of Fisher criterion.

\subsection{Related Methods}
Most existing person re-identification methods try to build robust feature descriptors and learn discriminative distance metrics. For feature descriptors, several works have been proposed to capture the invariant and discriminative properties of human images \cite{SDALF,LOMO,GOG,ELF,ColorInvariants,midlevel,colornames,LisantiPAMI14}. Specifically, GOG\cite{GOG} and LOMO\cite{LOMO} descriptors have shown impressive robustness against illumination, pose and viewpoint changes.

For recognition purposes, many metric learning methods have been proposed recently \cite{PRDC,RPLM,LFDA:CVPR,rPcca,KISSME,LOMO,ExPolyFeatMap,Zheng:nfst,SSSVM}. Most of the metric learning methods in re-ID originated elsewhere and are applied with suitable modification for overcoming the additional challenges in re-identification. Köstinger et al. proposed an efficient metric called KISSME \cite{KISSME} using log likelihood ratio test of two Gaussian distributions. Hirzer et al. \cite{RPLM} used a relaxed positive semi definite constraint of the Mahalanobis metric. Zheng et al. proposed PRDC\cite{PRDC} where the  metric is learned to maximize the probability of a pair of true match having a smaller distance than that of a wrong match pair.
As an improvement for KISSME\cite{KISSME}, Liao et al. proposed XQDA\cite{LOMO} to learn a more discriminative distance metric and a low-dimensional subspace simultaneously. In \cite{LFDA:CVPR}, Pedagadi et al. successfully applied  Local Fisher Discriminant Analysis (LFDA) \cite{LFDA:ICML} which is a variant of Fisher discriminant analysis to preserve the local structure.

Most metric learning methods based on Fisher-type criterion suffer from the small sample size (SSS) problem \cite{Zheng:nfst,guo:nfst}. The dimensionality of various efficient feature descriptors like LOMO\cite{LOMO} and GOG\cite{GOG} are in ten thousands and too high compared to the number of samples typically available for training. This makes the within class scatter matrix singular. Some methods use matrix regularization  \cite{LFDA:CVPR,rPcca,LOMO,LisantiDSC,GOG} or unsupervised dimensionality reduction \cite{KISSME,LFDA:CVPR} to overcome the singularity which makes them less discriminative and suboptimal. Also these methods typically  have a number of free parameters to tune. 

Recently, Null Foley-Sammon Transform (NFST)\cite{Zheng:nfst,bodesheim:novelty,guo:nfst} has gained increasing attention in computer vision applications. NFST was proposed in \cite{Zheng:nfst} to address the SSS problem in re-ID. They find a transformation which collapses the intra class training samples into a single point. By restricting the between class variance to be non zero, they maximize the Fisher discriminant criterion without the need of using any regularization or unsupervised dimensionality reduction. 

In this paper, we first identify a serious limitation of NFST, i.e. though NFST minimizes the intra-class distance to zero for all training data, it fails to maximize the inter class distance and has serious consequences creating suboptimality in generalizing the discrimination for test data samples when the test sample does not map to the corresponding singular points. Secondly, we propose a novel \color{black}metric learning framework called Nullspace Kernel Maximum Margin Metric Learning (NK3ML). The method learns a discriminative metric subspace to maximize the inter-class distance as well as minimize the intra-class distance to zero. NK3ML efficiently addresses the suboptimality of NFST in generalizing the discrimination to test data samples also. In particular, NK3ML first take advantage of NFST to find a low dimensional discriminative nullspace to collapse the intra class samples into a single point. Later NK3ML utilizes a secondary metric learning framework 
to learn a discriminant subspace using the nullspace to maximally separate the inter-class distance. 
NK3ML also uses a non-linear mapping of the discriminative nullspace into an infinite dimensional space using an appropriate kernel to further increase the maximum attainable margin between the inter class samples. The proposed NK3ML does not require regularization nor unsupervised dimensionality reduction and efficiently addresses the SSS problem as well as the suboptimality of NFST in generalizing the discrimination for test data samples. The proposed NK3ML has a closed from solution and has no free parameters to tune. 

We first explain NFST in Section \ref{sec:NFST}. Later we present NK3ML in Section \ref{sec:NK3ML} and the experimental results in Section \ref{sec:Exp}.

\section{Null Foley-Sammon Transform}
\label{sec:NFST}

\subsection{Foley-Sammon Transform}
\label{ssec:FST}
The objective of Foley-Sammon Transform (FST) \cite{sammon:odp,okada:oosda} is to learn optimal discriminant vectors $\mathbf{w} \in \mathbb{R}^{d}$ that maximize the \textit{Fisher criterion} $J_F(\mathbf{w})$ under orthonormal constraints:
\begin{equation}
\label{eqn:FC}
J_F(\mathbf{w})  = \dfrac{\mathbf{w}^T \mathbf{S}_b \mathbf{w}}{\mathbf{w}^T \mathbf{S}_w \mathbf{w}} \,.
\end{equation}
$\mathbf{S}_w$ represents the \textit{within class scatter matrix} and $\mathbf{S}_b$ the \textit{between class scatter matrix}. $\mathbf{x}\in \mathbb{R}^d$ are the data samples with classes $\mathcal{C}_1,\ldots,\mathcal{C}_c$ where $c$ is the total number of classes. Let $n$ be the total number of samples and $n_i$ the number of samples in class $\mathcal{C}_i$. 
FST tries to maximize the between class distance and minimize the within class distance simultaneously by maximizing the Fisher criterion.

The optimal discriminant vectors of FST are generated using the following steps. The first discriminant vector $\mathbf{w}_{1}$ of FST is the unit vector that maximizes $J_F(\mathbf{w}_1)$. 
\color{black} If $\mathbf{S}_w$ is nonsingular, the solution becomes a conventional eigenvalue problem: $\mathbf{S}_w^{-1} \mathbf{S}_b \mathbf{w} = \lambda  \mathbf{w}$, and can be solved by the normalized eigenvector of $\mathbf{S}_w^{-1} \mathbf{S}_b$ corresponding to its largest eigenvalue. 
 The \textit{i}th discriminant vector $\mathbf{w}_{i}$ of FST is calculated by the following optimization problem with orthonormality constraints:
\begin{equation}
\begin{aligned}
& \underset{||\mathbf{w}_i|| = 1,\mathbf{w}_{i}^T \mathbf{w}_{j} = 0}   {\text{maximize}} & & \{J_F(\mathbf{w}_i)\} \quad j = 1, \ldots, i-1 \,.
\label{eqn:OptFST}
\end{aligned}
\end{equation}
A major drawback of FST is that it cannot be directly applied when $\mathbf{S}_w$ becomes singular in small sample size (SSS) problems. The SSS problem occures when $n<d$.
Common solutions include adding regularization term to $\mathbf{S}_w$ or reducing the dimensionality using PCA, which makes them suboptimal.

\subsection{Null Foley-Sammon Transform}
\label{ssec:NFST}
The suboptimality due to SSS problem in FST is overcome in an efficient way using Null Foley-Sammon Transform (NFST). The objective of NFST is to find orthonormal discriminant vectors satisfying the following set of constraints: 
\begin{equation}
\label{eqn:NFSTC1}
\mathbf{w}^T \mathbf{S}_w \mathbf{w} = 0, \;\;\;\;
\mathbf{w}^T \mathbf{S}_b \mathbf{w} > 0 \,.
\end{equation}
Each discriminant vector $\mathbf{w}$ should satisfy zero within-class scatter and positive between-class scatter. This leads to $J_F(\mathbf{w}) \to \infty$ and thus NFST tries to attain the best separability in terms of Fisher criterion. Such a vector $\mathbf{w}$ is called \textit{Null Projecting Direction} (NPD). 
The zero within-class scatter ensures that the transformation using NPDs collapse the intra-class training samples into a single point.

\setlength{\parskip}{1.5mm}
\noindent \textbf{Obtaining Null Projecting Directions}: We explain how to obtain the Null Projecting Direction (NPD) of NFST. The total class scatter matrix $\mathbf{S}_t$ is defined as $\mathbf{S}_t = \mathbf{S}_b + \mathbf{S}_w $. We also have $\mathbf{S}_t = \frac{1}{n}\mathbf{P}_t \mathbf{P}_t^T$, where $\mathbf{P}_t$ consists of zero mean data $\mathbf{x}_{1}-\mathbf{m},\ldots, \mathbf{x}_{n}-\mathbf{m}$ as its columns.
Let $\mathbf{Z}_t$ and $\mathbf{Z}_w$ be the null space of $\mathbf{S}_t$ and $\mathbf{S}_w$ respectively. Let $\mathbf{Z}^\perp_t$ represent orthogonal complement of $\mathbf{Z}_t$. Note the lemmas\cite{guo:nfst}.
\setlength{\parskip}{0mm}

\textit{Lemma 1:} Let $\mathbf{A}$ be a positive semidefinite matrix. Then $\mathbf{w}^TA\mathbf{w}=0$ iff $\mathbf{A}\mathbf{w}=0$. 

\textit{Lemma 2:} If $\mathbf{w}$ is an NPD, then $\mathbf{w} \in (\mathbf{Z}_t^\perp \cap \mathbf{Z}_w)$.

\textit{Lemma 3:} For small sample size (SSS) case, there exists exactly $c-1$ NPDs, $c$ being the number of classes.

In order to obtain the NPDs, we first obtain vectors from the space $\mathbf{Z}_t^\perp$. From this space, we next obtain vectors that also satisfy $\mathbf{w} \in \mathbf{Z}_w$. A set of orthonormal vectors can be obtained from the resultant vectors which form the NPDs. 

Based on the lemmas, $\mathbf{Z}_t$ can be solved as:
\begin{eqnarray}
\begin{aligned}
\mathbf{Z}_t  &= \lbrace  \mathbf{w}  \; \vert \;  \mathbf{S}_t \mathbf{w} = 0\rbrace = \lbrace  \mathbf{w} \; \vert \;  \mathbf{w}^T \mathbf{S}_t \mathbf{w} = 0\rbrace \\
&= \lbrace  \mathbf{w} \; \vert \;  (\mathbf{P}_t^T \mathbf{w})^T (\mathbf{P}_t^T \mathbf{w})  = 0\rbrace = \lbrace  \mathbf{w} \; \vert \;  \mathbf{P}_t^T \mathbf{w}  = 0\rbrace \,.
\end{aligned}
\end{eqnarray}
Thus $\mathbf{Z}_t$ is the null space of $\mathbf{P}_t^T$. So $\mathbf{Z}^\perp_t$ is the row space of $\mathbf{P}_t^T$, which is the column space of $\mathbf{P}_t$. Therefore $\mathbf{Z}^\perp_t$ is the subspace spanned by zero mean data. $\mathbf{Z}^\perp_t$ can be represented using an orthonormal basis $\mathbf{Q} = (\theta_1,\ldots, \theta_{n-1})$, where $n$ is the total number of samples. The basis $\mathbf{Q}$ can be obtained using Gram-Schmidt orthonormalization procedure. Any vector in $\mathbf{Z}^\perp_t$ can hence be represented as: 
\begin{equation}
\label{eqn:beta}
\mathbf{w} = \beta_1 \theta_1 + \ldots + \beta_{n-1} \theta_{n-1} = \mathbf{Q}\bm{\beta}\,.
\end{equation}
A vector $\mathbf{w}$, satisfying Eqn. (\ref{eqn:beta}) for any $\bm{\beta}$, belongs to $\mathbf{Z}_t^\perp$.  Now we have to find those specific  $\bm{\beta}$ which ensures $\mathbf{w}\in \mathbf{Z}_w$. They can be found by substituting (\ref{eqn:beta}) in the condition for $\mathbf{w}\in \mathbf{Z}_w$ as follows:
\begin{eqnarray}
\begin{aligned}
0 &= \mathbf{S}_w \mathbf{w}  = \mathbf{w}^T \mathbf{S}_w \mathbf{w} = (\mathbf{Q\beta})^T \mathbf{S}_w (\mathbf{Q}\bm{\beta})\\
&= \bm{\beta}^T(\mathbf{Q}^T \mathbf{S}_w \mathbf{Q})\bm{\beta} = \mathbf{Q}^T \mathbf{S}_w \mathbf{Q}\bm{\beta}\,.
\end{aligned}
\end{eqnarray}
Hence $\bm{\beta}$ can be solved by finding the null space of $\mathbf{Q}^T \mathbf{S}_w \mathbf{Q}$. The set of solutions $\{\bm{\beta}\}$ can be chosen orthonormal. Since the dimension of $\mathbf{w} \in (\mathbf{Z}_t^\perp \cap \mathbf{Z}_w)$ is $c-1$ \cite{guo:nfst}, we get $c-1$ solutions for $\bm{\beta}$.  The $c-1$ NPDs can now be computed using (\ref{eqn:beta}). Since $\mathbf{Q}$ and $\{\bm{\beta}\}$ are orthonormal, the resulting NPDs are also orthonormal. The projection matrix $\mathbf{W}_N \in \mathbb{R}^{d \times (c-1)}$ of NFST now constitutes of the $c-1$ NPDs as its columns.

\section{Nullspace Kernel Maximum Margin Metric Learning}
\label{sec:NK3ML}
\begin{figure*}[t!]
\begin{center}
   \includegraphics[width=0.7\linewidth]{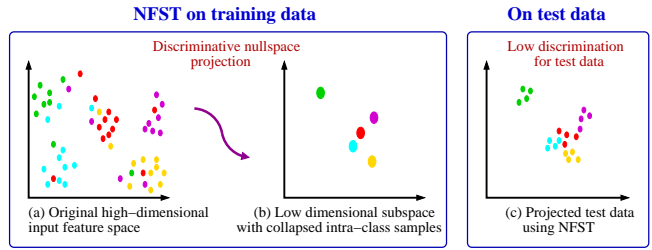}
\end{center}
   \caption{Illustration of the suboptimality in NFST. Each color corresponds to distinct classes.}
\label{fig:N3ML1}
\end{figure*}

Methods based on Fisher criterion, in general, learn the discriminant vectors using the training samples so that the vectors generalize well for the test data also in terms of separability of classes. NFST\cite{guo:nfst,bodesheim:novelty} was proposed in \cite{Zheng:nfst} to address the SSS problem in re-ID. They find a transformation by collapsing the intra-class samples into a single point. We identify a serious limitation of NFST. Maximizing $J_F(\mathbf{w})$ in Eqn. (\ref{eqn:FC}) by making the dinominator to zero, does not allow to make use of the information contained in the numerator. As illustrated in Fig. \ref{fig:N3ML1},  the mapped singular points in the NFST projected space for two different classes may be quite close. Thus, when a test data is projected into this NFST nullspace, it no longer maps to the same singular point. Rather, it maps to a point close to the above point. But this projected point may be closer to the singular point for the other class and misclassification takes place. Under the NFST formulation, one has no control on this aspect as one makes $\mathbf{w}^T\mathbf{S}_w \mathbf{w} = 0$, but $\mathbf{w}^T\mathbf{S}_b \mathbf{w}$ may also be very small instead of being large, and the classification performance may be very poor. 

\begin{figure*}[t]
\begin{center}
   \includegraphics[width=1\linewidth]{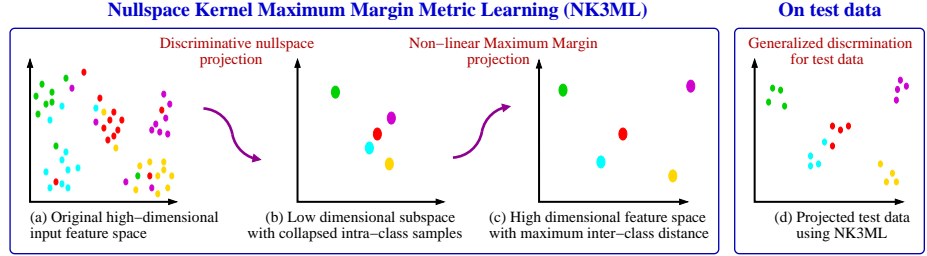}
\end{center}
   \caption{Illustration of our method NK3ML. Each color corresponds to distinct classes.}
\label{fig:N3ML2}
\end{figure*}

In this paper we propose a metric learning framework, namely, Nullspace Kernel Maximum Margin Metric Learning (NK3ML) to improve the limitation of NFST and better  handle the classification of high dimensional data. As shown in Fig. \ref{fig:N3ML2}, NK3ML first take advantage of NFST to find a low dimensional discriminative nullspace to collapse the intra-class samples into a single point. Later it uses a modified version of Maximum Margin Criterion (MMC)\cite{haifeng:mmc} to learn a discriminant subspace using the nullspace to maximally separate the inter-class distance.
Further, to obtain the benefit of kernel based techniques, instead of using the MMC, we obtain the                                                                                                                                                                                                                                                                                                                                                                                                                                                                                                                                                                                                                                                                                                                                                                                                                                                                                                                                                                                                                                                                                                                                                                                                                                                                                                                                                                                                                                                                                                                                                                                                                                                                                                                                                                                                                                                                                                                                                                                                                                                                                                                                                                                                                                       \textit{Normalized Kernel Maximum margin criterion} (NKMMC) which is efficient and robust to learn the discriminant subspace to maximize the distances among the classes. NK3ML can efficiently address the suboptimality of NFST in enhancing the discrimination to test data samples also.

\subsection{Maximum Margin Criterion}
\label{ssec:MMC}
\textit{Maximum margin criterion} (MMC) \cite{haifeng:mmc,mmc:pami} is an efficient way to learn a discriminant subspace which maximize the distances between classes. 
For the separability of classes $\mathcal{C}_1,\ldots,\mathcal{C}_c$, the maximum margin criterion is defined as 
\begin{equation}
\label{eqn:MMC1}
J  = \frac{1}{2} \sum\limits_{i=1}^{c} \sum\limits_{j=1}^{c}  \textit{p}_i \textit{p}_j d(\mathcal{C}_i,\mathcal{C}_j) \,,
\end{equation}
where the inter-class margin (or distance) of class $\mathcal{C}_i$ and $\mathcal{C}_j$ is defined as 
\begin{equation}
\label{eqn:MMC2}
d(\mathcal{C}_i,\mathcal{C}_j) = d(\mathbf{m}_i,\mathbf{m}_j) - s(\mathcal{C}_i) - s(\mathcal{C}_j) \,,
\end{equation}
and $d(\mathbf{m}_i,\mathbf{m}_j)$ represents the squared Euclidean distance between mean vectors $\mathbf{m}_i$ and $\mathbf{m}_j$ of classes $\mathcal{C}_i$ and $\mathcal{C}_j$, respectively. $s(\mathcal{C}_i)$ is the scatter of class $\mathcal{C}_i$, estimated as $s(\mathcal{C}_i) = tr(\mathbf{S}_i)$ where $\mathbf{S}_i$ is the within class scatter matrix of class $\mathcal{C}_i$. The inter-class margin can be solved to get $d(\mathcal{C}_i,\mathcal{C}_j)  = \textit{tr} \;  (\mathbf{S}_b  -  \mathbf{S}_w)$.
A set of $r$ unit linear discriminant vectors $\{\mathbf{v}_k \in \mathbb{R}^{d} | k=1,\ldots,r\}$ is learned such that they maximize $J$ in the projected subspace.
If $\mathbf{V} \in \mathbb{R}^{d \times r}$ is the projection matrix, the MMC criterion becomes $J(\mathbf{V})  = \textit{tr} \;  (\mathbf{V}^T (\mathbf{S}_b  -  \mathbf{S}_w) \mathbf{V})$. The optimization problem can be equivalently written as:
\begin{equation}
\begin{aligned}
& \underset{\mathbf{v}_{k}}{\text{maximize}} & &\sum\limits_{k=1}^r \;  \mathbf{v}^T_{k} (\mathbf{S}_b  -  \mathbf{S}_w) \mathbf{v}_{k} \,,\\
\label{eqn:OptMMC}
& \text{subject to}     & & \mathbf{v}^T_{k} \mathbf{v}_{k} = 1 \,, \qquad k=1,\ldots,r \ .
\end{aligned} 
\end{equation}
The optimal solutions are obtained by finding the normalized eigenvectors of $\mathbf{S}_b - \mathbf{S}_w$ corresponding to its first $r$ largest eigenvectors.

\subsection{Kernel Maximum Margin Criterion}
\label{ssec:KMMC}
Kernels methods are well known techniques to learn non-linear discriminant vectors. They use an appropriate non-linear function $\Phi(\mathbf{z})$ to map the input data $\mathbf{z}$ to a higher dimensional feature space $\mathcal{F}$ and find discriminant vectors $\mathbf{v}_k \in \mathcal{F}$. Given $n$ training data samples and a kernel function $k(\mathbf{z}_i, \mathbf{z}_j) = \langle \mathbf{\Phi}(\mathbf{z}_i), \mathbf{\Phi}(\mathbf{z}_j) \rangle $, we can calculate the kernel matrix $\mathbf{K}\in \mathbb{R}^{n \times n}$. The matrix $\mathbf{K}_i \in \mathbb{R}^{n \times n_i}$
for the \textit{i}th class with $n_i$ samples is $(\mathbf{K}_i)_{pq}:=k(\mathbf{z}_p,\mathbf{z}_q^{(i)})$. As every discriminant vector $\mathbf{v}_k$ lies in the span of the mapped data samples, it can be expressed in the form $\mathbf{v}_k = \sum _{j=1}^{n} (\bm{\alpha}_{k})_j \mathbf{\Phi}(\mathbf{z}_j)$,
where $(\bm{\alpha}_{k})_j$ is the \textit{j}th element of the vector $\bm{\alpha}_{k} \in \mathbb{R}^{n}$, which constitutes the expansion coefficients of $\mathbf{v}_k$. 
The optimization problem proposed for Kernel Maximum Margin Criterion (KMMC)\cite{haifeng:mmc} is:
\begin{equation}
\begin{aligned}
& \underset{\bm{\alpha}_{k}}{\text{maximize}} & & \sum\limits_{k=1}^r \;  \bm{\alpha}^T_{k} (\mathbf{M}  -  \mathbf{N}) \bm{\alpha}_{k} \,,\\
\label{eqn:KMMCFinal}
& \text{subject to}     & & \bm{\alpha}^T_{k} \bm{\alpha}_{k} = 1 \,,
\end{aligned}
\end{equation}
where $\mathbf{N}:= \sum_{i = 1}^{c} \frac{1}{n} \mathbf{K}_i(\mathbf{I}_{n_i}- \frac{1}{n_i} \mathbf{1}_{n_i}\mathbf{1}_{n_i}^T)\mathbf{K}_i^T$, 
$\;\; \mathbf{I}_{n_i}$ is $(n_i \times n_i)$ identity matrix; $\mathbf{1}_{n_i}$ is $n_i$ dimensional vector of ones and
$\mathbf{M}=\sum_{i = 1}^{c} \frac{1}{n_i} (\widetilde{\mathbf{m}}_i-\widetilde{\mathbf{m}})(\widetilde{\mathbf{m}}_i- \widetilde{\mathbf{m}})^T$;
$\;\;\widetilde{\mathbf{m}} := \frac{1}{n} \sum_{i=1}^{c} n_i \widetilde{\mathbf{m}}_i$ and $(\widetilde{\mathbf{m}}_i)_j := \frac{1}{n_i}  \sum_{\mathbf{z} \in \mathcal{C}_i}    k(\mathbf{z},\mathbf{z}_j)$.
 The optimal solutions are the normalized eigenvectors of ${(\mathbf{M} - \mathbf{N})}$, corresponding to its first $r$ largest eigenvalues.

\subsection{NK3ML}
\label{ssec:NK3ML}

The kernalized optimization problem given in (\ref{eqn:KMMCFinal}) obtained by KMMC\cite{haifeng:mmc} does not enforce normalization of discriminant vectors in the feature space, but rather uses normalization constraint on eigenvector expansion coefficient vector $\bm{\alpha}_k$. In NK3ML, we require the discriminant vectors obtained by KMMC  to be normalized, i.e., $\mathbf{v}_k^T\mathbf{v}_k =1$. The normalized discriminant vectors are important to preserve the shape of the distribution of data. Hence we derive \textit{Normalized Kernel Maximum Margin Criterion} (NKMMC) as follows. We rewrite the discriminant vector $\mathbf{v}_k$ as:
\begin{eqnarray}
\begin{aligned}
\mathbf{v}_k &= \sum \limits_{j=1}^{n} (\bm{\alpha}_{k})_j \mathbf{\Phi}(\mathbf{z}_j) = \Big[\mathbf{\Phi}(\mathbf{z}_1) \; \mathbf{\Phi}(\mathbf{z}_2)\; \ldots \;\mathbf{\Phi}(\mathbf{z}_n) \Big] \bm{\alpha}_k \,.
\end{aligned}
\end{eqnarray}
\noindent Then normalization constraint becomes
\begin{eqnarray}
&\Big(\sum \limits_{j=1}^{n} (\bm{\alpha}_{k})_j \mathbf{\Phi}(\mathbf{z}_j)\Big)^T \Big(\sum \limits_{j=1}^{n} (\bm{\alpha}_{k})_j \mathbf{\Phi}(\mathbf{z}_j)\Big) =1 \nonumber\\
\Rightarrow &\bm{\alpha}_k^T \mathbf{K} \bm{\alpha}_k =1 \,.
\end{eqnarray}
where $\mathbf{K}$ is the kernel matrix. The optimization problem in (\ref{eqn:KMMCFinal}) can now be reformulated to enforce normalized discriminant vectors as follows.
\begin{equation}
\begin{aligned}
& \underset{\bm{\alpha}_{k}}{\text{maximize}} & & \sum\limits_{k=1}^r \;  \bm{\alpha}^T_{k} (\mathbf{M}  -  \mathbf{N}) \bm{\alpha}_{k} \,,\\
\label{eqn:CKMMCFinal}
& \text{subject to}     & & \bm{\alpha}^T_{k}\mathbf{K} \bm{\alpha}_{k} = 1 \,.
\end{aligned}
\end{equation}
\noindent We introduce a Lagrangian to solve the above problem.
\begin{equation}
\mathcal{L}(\bm{\alpha}_k,\lambda_k) = 
\sum \limits_{k=1}^{r} \bm{\alpha}_k^T (\mathbf{M-N}) \bm{\alpha}_k + \lambda_k(\bm{\alpha}^T_{k}\mathbf{K} \bm{\alpha}_{k} - 1) \,,
\end{equation}
where $\lambda_k$ is the Lagrangian multiplier. The Lagrangian $\mathcal{L}$ has to be maximized with respect to $\bm{\alpha}_{k}$ and the multipliers $\lambda_k$. The derivatives of $\mathcal{L}$ with respect to $\bm{\alpha}_k$ should vanish at the stationary point.
\begin{eqnarray}
\begin{aligned}
\frac{\partial \mathcal{L}(\bm{\alpha}_k,\lambda_k)}{\partial \bm{\alpha}_k} &= (\mathbf{M-N} - \lambda_k \mathbf{K})\bm{\alpha}_k =0 
 \; \;\;\;\forall \; k=1,\ldots, r\\
 & \Rightarrow (\mathbf{M-N})\bm{\alpha}_k = \lambda_k \mathbf{K} \bm{\alpha}_k  \,.
 \end{aligned}
\end{eqnarray}
This is a generalized eigenvalue problem. $\lambda_k$'s are the generalized eigenvalues and $\bm{\alpha}_k$'s the generalized eigenvectors of ($\mathbf{M-N}$) and $\mathbf{K}$. The objective function at this stationary point is given as:
\begin{equation}
\sum\limits_{k=1}^r \;  \bm{\alpha}^T_{k} (\mathbf{M}  -  \mathbf{N})\bm{\alpha}_{k} = \sum\limits_{k=1}^r \lambda_k \bm{\alpha}^T_{k}\mathbf{K}\bm{\alpha}_{k} = \sum\limits_{k=1}^r \lambda_k \,.
\end{equation}
\noindent Hence the objective function in NKMMC is maximized by the \textit{generalized} eigenvectors corresponding to the first $r$ generalized eigenvalues of ($\mathbf{M-N}$) and $\mathbf{K}$. We choose all the eigenvectors with positive eigenvalues, since they ensure maximum inter-class margin, i.e., the samples of different classes are well separated in the direction of these eigenvectors. It should be noted that our  NKMMC has a different solution from that of original KMMC\cite{haifeng:mmc}, since KMMC uses standard eigenvectors of $\mathbf{M-N}$.

NFST is first used to  learn the  discriminant vectors using the training data \{$\mathbf{x}\}$. The discriminants of NFST form the projection matrix $\mathbf{W}_{N}$. Each training data sample $\mathbf{x}\in \mathbb{R}^d$ is projected as
\begin{equation}
\mathbf{z} = \mathbf{W}^T_{N} \mathbf{x} \,.
\end{equation}
Each projected data sample $\mathbf{z}\in \mathbb{R}^{c-1}$ now lies in the discriminative nullspace of NFST.  Now we use all the projected data $\{\mathbf{z}\}$ for learning the secondary distance metric using NKMMC.

Any general feature vector $\widetilde{\mathbf{x}} \in \mathbb{R}^d$ can be projected onto the discriminant vector $\mathbf{v}_k$ of NK3ML in two steps:\\
\textit{Step 1}: Project $\widetilde{\mathbf{x}}$ onto the nullspace of NFST to get $\widetilde{\mathbf{z}}$ : 
\begin{eqnarray}
\widetilde{\mathbf{z}} = \mathbf{W}^T_{N} \widetilde{\mathbf{x}} \,.
\end{eqnarray}
\textit{Step 2}: Project the $\widetilde{\mathbf{z}} $ onto the discriminant vector $\mathbf{v}_k$ of NKMMC:
\begin{eqnarray}
\mathbf{v}_k^T \Phi(\widetilde{\mathbf{z}}) &=& \Big( \sum\limits_{j = 1}^{n} (\bm{\alpha}_{k})_j \mathbf{\Phi}(\mathbf{z}_j)\Big)^T \Phi(\widetilde{\mathbf{z}})
= \sum\limits_{j = 1}^{n} (\bm{\alpha}_{k})_j  k(\mathbf{z}_j,\widetilde{\mathbf{z}}) \,.
\end{eqnarray}

The proposed NK3ML does not require any regularization or unsupervised dimensionality reduction and can efficiently address the SSS problem as well as the suboptimality of NFST in generalizing the discrimination for test data samples. The NK3ML has a closed form solution and no free parameters to tune. The only issue to be decided is what kernel to be used. In effect what the proposed method does is to project the data into the NFST nullspace, where the dimensionality of the feature space is reduced to force all points belonging to a given class to a single point. In the second stage, the dimensionality is increased by using an appropriate kernel in conjunction with NKMMC, thereby allowing us to enhance the between class distance. This provides a better margin while classifying the test samples.

\section{Experimental Results}
\label{sec:Exp}
\noindent\textbf{Parameter Settings}: There are no free parameters to tune in NK3ML, unlike most state-of-the-art methods which have to carefully tune their parameters to attain their best results. In all the experiments, we use the RBF kernel whose kernel width is set to be the root mean squared pairwise distance among the samples. 

\setlength{\parskip}{1mm}
\noindent\textbf{Datasets}: The proposed NK3ML is evaluated on four popular benchmark datasets: PRID450S\cite{PRID450S}, GRID\cite{GRID1}, CUHK01\cite{CUHK01} and VIPeR\cite{ELF}, respectively contains 450, 250, 971, and 632 identities captured in two disjoint camera views. 
CUHK01 contains two images for each person in one camera view and all other datasets contain just one image. Quite naturally, these datasets constitute the extreme examples of SSS. Following the conventional experimental setup \cite{song:scalableManifold,GOG,SCSP,LOMO,metric_ensembles,colornames}, each dataset is randomly divided into training and test sets, each having half of the identities. During testing, the probe images are matched against the gallery. In the test sets of all datasets, except GRID, the number of probe images and gallery images are equal. The test set of GRID has additional 775 gallery images that do not belong to the 250 identities. The procedure is repeated 10 times and the average rank scores are reported.

\noindent\textbf{Features}: Most existing methods use a fixed feature descriptor for all datasets. 
Such an approach is less efficient to represent the intrinsic characteristics of each dataset. Hence in NK3ML, we use specific set of feature descriptors for each dataset.
We choose from the standard feature descriptors GOG\cite{GOG} and WHOS\cite{LisantiPAMI14}. We also use an improved version of LOMO\cite{LOMO} descriptor, which we call \textit{LOMO*}. 
We generate it by concatenating the LOMO features generated using YUV and RGB color spaces separately.

\noindent\textbf{Method of Comparison}: We use only the available data in each dataset for training. No separate pre-processing of the features or images (such as domain adaptation / body parts detection), or post-processing of the classifier has been used in the study. There has been some efforts on 
using even the test data for re-ranking of re-ID results \cite{song:scalableManifold,Reranking:kreciprocal,SHaPE} to boost up the accuracy. But these techniques being not suitable for any real time applications, we refrain from using such supplementary methods in our proposal. 
\setlength{\parskip}{0mm}

\subsection{Comparison with Baselines}
\label{ssec:Baselines}

\begin{table}[ht]
\setlength{\abovecaptionskip}{10pt} 
\begin{minipage}[c]{0.59\linewidth}
\centering
    \caption{Comparison of NK3ML with baselines on \\GRID and PRID450S datasets}
\begin{tabular}[t]{|l||c|c||c|c|}
\hline
\multirow{2}{*}{Methods} & \multicolumn{2}{c||}{GRID} & \multicolumn{2}{c|}{PRID450S} \\ \cline{2-5} 
 & Rank1 & Rank10 & Rank1 & Rank10\\
\hline\hline
\small WHOS + \color{red}NK3ML  & \color{red}21.20  & \color{red}55.60  & \color{red}50.67  & \color{red}88.09 \\
WHOS + NFST  & 18.64 & 52.32 & 42.58 &  77.07 \\
WHOS + KNFST & 21.12   & 54.32 & 45.87 &  85.78  \\
WHOS + XQDA  & 18.72  & 52.56  & 43.38  & 77.91\\
\hline \hline
LOMO + \color{red}NK3ML & \color{red}18.24  & \color{red}43.76 & \color{red}60.62 & \color{red}91.96\\
LOMO + NFST  & 17.04 & 42.64 & 58.84 &  89.42\\
LOMO + KNFST & 14.88  & 41.28 & 59.47 &  91.96\\
LOMO + XQDA  & 16.56  & 41.84  & 59.78 & 90.09\\
\hline \hline
GOG + \color{red}NK3ML & \color{red}26.96& \color{red}57.52 & \color{red}68.04  & \color{red}95.07\\
GOG + NFST  & 24.88  & 58.00 & 67.60 & 94.18 \\
GOG + KNFST & 24.88 & 53.28 & 64.80 & 94.00\\
GOG + XQDA  & 24.80 & 58.40 & 68.00 &  94.36\\
\hline
\end{tabular}
    \label{table:GRID:prid450sBaseline}
\end{minipage}\hfill
\begin{minipage}[c]{0.4\linewidth}
\centering
\includegraphics[width=40mm]{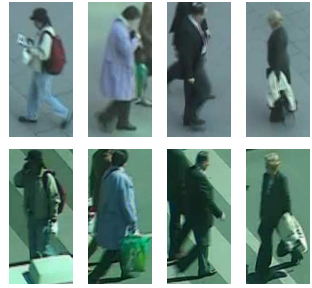}
\captionof{figure}{Sample images of PRID450S dataset. Images with the same column corresponds to the same identities.}
\label{fig:image}
\end{minipage}
\end{table}

In Table \ref{table:GRID:prid450sBaseline}, we compare the performances of NK3ML with the baseline metric learning methods. As NK3ML is proposed as an  improvement to address the limitations of NFST, we first compare the performance of NK3ML with NFST. For fair comparison with NFST, we also use its kernalized version KNFST\cite{Zheng:nfst}. KNFST is also the state-of-the-art metric learning method applied for LOMO descriptor. For uniformity, all metric learning methods are evaluated using the same standard feature descriptors LOMO\cite{LOMO}, WHOS\cite{LisantiPAMI14} and GOG\cite{GOG}.
 We also compare with Cross-view Quadratic Discriminant Analysis (XQDA)\cite{GOG} which is the state of the art metric learning method for GOG descriptor. XQDA is also successfully applied with LOMO in many cases\cite{LOMO}. We use GRID and PRID450S datasets for comparison with the baselines. GRID is a pretty difficult person re-identification dataset having poor image quality with large variations in pose and illuminations, which makes it very challenging to obtain good matching accuracies. PRID450S is also a challenging dataset due to the partial occlusion, background interference and viewpoint changes. From the results in Table \ref{table:GRID:prid450sBaseline}, it can be seen that NK3ML provides significant performance gains against all the baselines for all the standard feature descriptors.

\setlength{\parskip}{0.5mm}
\noindent\textbf{Comparison with NFST}: NK3ML provides a good performance gain against NFST. In particular for PRID450S dataset, when compared using WHOS, NK3ML provides an improvement of 8.09\% at rank-1 and 11.02\% at rank-10.
Similar gain can also be seen while using LOMO and GOG features for both GRID and PRID450S datasets.

\noindent\textbf{Comparison with KNFST}: In spite of KNFST being the state-of-the-art metric learning method for LOMO descriptor, NK3ML outperforms KNFST with a significant difference. In GRID dataset, NK3ML gains 3.36\% in rank-1
and 2.48\% in rank-10. 
Similar improvements are seen for other features also for both datasets.

\noindent\textbf{Comparison with XQDA}: For GOG descriptor, XQDA is the state of the art metric learning method. At rank-1, NK3ML gains 2.16\% in GRID.
Similarly, it gains 7.29\% at rank-1 in PRID450S using WHOS descriptor.

\setlength{\parskip}{0mm}
Based on the above comparisons, it may be concluded that NK3ML attains a much better margin over NFST as expected from the theory.  Also NK3ML outperforms KNFST and XQDA for all aforementioned standard feature descriptors.

\setlength{\textfloatsep}{5mm}
\begin{table}[t]
\caption{Comparison with state-of-the-art results on (a) GRID and (b) PRID450S dataset. The best and second best scores are shown in red and blue, respectively. The methods with a * signifies pre/post-processing based methods}
\begin{center}
\subfloat[Subtable 1 list of tables text][GRID dataset]{
\begin{tabular}[c]{|l|c|c|c|}
\hline
Methods &  Rank1 & Rank10 & Rank20 \\
\hline\hline
\small MtMCML\cite{MtMCML}		&	14.08	&	45.84	&	59.84	\\
KNFST\cite{Zheng:nfst}   & 14.88	&	41.28	&	50.88 \\
PolyMap\cite{ExPolyFeatMap}		&	16.30	&	46.00	&	57.60	\\
LOMO+XQDA\cite{LOMO}		&	16.56	&	41.84	&	52.40	\\
MLAPG\cite{MLAPG}		&	16.64	&	41.20	&	52.96	\\
KEPLER\cite{KEPLER}		&	18.40	&	50.24	&	61.44	\\
DR-KISS\cite{DR-KISS}		&	20.60	&	51.40	&	62.60	\\
SSSVM\cite{SSSVM}  & 22.40	& 51.28	&	61.20\\
SCSP\cite{SCSP}		&	24.24	&	54.08	&	65.20	\\
GOG+XQDA\cite{GOG}	&	\color{blue}\textbf{24.80}	&	\color{blue}\textbf{58.40}	&	\color{blue}\textbf{68.88}	\\
NK3ML(Ours) &  \color{red}\textbf{27.20}	& \color{red}\textbf{60.96} & \color{red}\textbf{71.04}\\
\hline \hline
*SSDAL\cite{SSDAL}		&	22.40	&	48.00	&	58.40	\\
*SSM\cite{song:scalableManifold}  & 27.20	&	\color{red}\textbf{61.12}	&	70.56\\
*OL-MANS\cite{OnlineNegSamples}  & \color{red}\textbf{30.16} &	49.20 &	59.36\\
\hline
\end{tabular}
\label{table:GRIDall}
}
\subfloat[Subtable 2 list of tables text][PRID450S dataset]{
\begin{tabular}[b]{|l|c|c|c|}
\hline
Methods  & Rank1 & Rank10 & Rank20 \\
\hline\hline
WARCA\cite{WARCA} & 24.58 & - & -\\
SCNCD\cite{colornames}		&	41.60	&	79.40	&	87.80	\\
CSL\cite{CSL}		&	44.40	&	82.20	&	89.80	\\
TMA\cite{TMA}		&	52.89	&	85.78	&	93.33	\\
KNFST\cite{Zheng:nfst} &	 59.47	&	91.96	&	96.53\\
LOMO+XQDA\cite{LOMO}		&	59.78	&	90.09	&	95.29	\\
SSSVM\cite{SSSVM}		&	60.49	&	88.58	&	93.60	\\
GOG+XQDA\cite{GOG}		&	\color{blue}\textbf{68.00} 	&	\color{blue}\textbf{94.36}	&	\color{blue}\textbf{97.64}	\\
NK3ML(Ours) & \color{red}\textbf{73.42}	& \color{red}\textbf{96.31}	& \color{red}\textbf{98.58}\\
\hline \hline
*Semantic\cite{Symantic}		&	44.90	&	77.50	&	86.70	\\
*SSM\cite{song:scalableManifold}	&	72.98	&	\color{red}\textbf{96.76}	&	\color{red}\textbf{99.11}	\\
\hline
\end{tabular}
\label{table:PRID450Sall}
}
\end{center}
\end{table}

\subsection{Comparison with State-of-the-art}
In the performance comparison of NK3ML with the state-of-the-art methods, \color{black} we also report the accuracies of pre/post processing methods on separate rows for completeness. As mentioned previously, direct comparisons of our results with pre/post processing methods are not advisable. However, even if such a comparison is made, we still have accuracies that are best or comparable to the best existing techniques on most of the evaluated datasets. Moreover, our approach is general enough to be easily integrated with the existing pre/post processing methods to further increase their accuracy.

\setlength{\parskip}{0.5mm}
\noindent \textbf{Experiments on GRID dataset:} We use GOG and LOMO* as the feature descriptor for GRID. 
Table \ref{table:GRIDall} shows the performance comparison of NK3ML. 
 GOG + XQDA\cite{GOG} reports the best performance of 24.8\% at rank-1 till date. NK3ML achieves an accuracy of 27.20\% at rank-1, outperforming GOG+XQDA by 2.40\%. 
At rank-1, NK3ML also outperforms all the post processing methods except OL-MANS\cite{OnlineNegSamples}, which uses the test data and train data together to learn a better similarity function. However, the penalty for misclassification at rank-1, if any,  severely affects the rank-N performance for OL-MANS. NK3ML outperforms OL-MANS by 11.76\% at rank-10 and 11.68\% at rank-20.

\setlength{\parskip}{1mm}
\noindent\textbf{Experiments on PRID450S dataset: }
GOG and LOMO* are used as the feature descriptor for PRID450S. 
NK3ML provides the best performances at all ranks, as shown in Table \ref{table:PRID450Sall}.  Especially, it provides an improvement margin of 5.42\% in rank-1 compared to the second best method GOG+XQDA\cite{GOG}.  At rank-1, NK3ML also outperforms all the post processing based methods. SSM\cite{song:scalableManifold} incorporates XQDA as the metric learning method. As analyzed in Section \ref{ssec:Baselines}, since NK3ML outperforms XQDA, it can be anticipated that even the re-ranking methods like SSM can benefit from NK3ML.\color{black}
\setlength{\parskip}{0mm}

\begin{table}[t]
\caption{Comparison with state-of-the-art results on CUHK01 dataset using (a) single-shot and (b) multi-shot settings. ** corresponds to deep learning based methods}
\begin{center}
\subfloat[Subtable 1 list of tables text][single-shot]{
\begin{tabular}{|l|c|c|c|}
\hline
Methods &  Rank1 & Rank10 & Rank20 \\
\hline\hline
\small 
MLFL\cite{midlevel}  & 34.30& 65.00&75.00\\
LOMO+XQDA\cite{LOMO}  & 50.00		&83.40		&89.51\\
KNFST\cite{Zheng:nfst}  &52.80 	&84.97		&91.07\\
CAMEL\cite{CAMEL}  & 57.30 &-	&-\\
GOG+XQDA\cite{GOG}	 & \color{blue}\textbf{57.89} &  \color{blue}\textbf{86.25} &  \color{blue}\textbf{92.14}\\
WARCA\cite{WARCA} & 58.34 &- &-\\
NK3ML(Ours) & \color{red}\textbf{67.09}& 	 \color{red}\textbf{91.85}&	 \color{red}\textbf{95.92}\\
\hline\hline
*Semantic\cite{Symantic}		&	32.70& 64.40 &76.30	\\
*MetricEnsemble\cite{metric_ensembles}  & 53.40 &  84.40&   90.50\\
**TPC\cite{TPC}  &53.70		&91.00&96.30\\
**Quadruplet\cite{Beyond:triplet_loss}  & 62.55& 89.71		&-\\
*DLPAR\cite{DLPAR} & \color{red}\textbf{72.30}		& \color{red}\textbf{94.90}&	\color{red}\textbf{97.20}\\
\hline
\end{tabular}
\label{table:CUHKM1}
}
\subfloat[Subtable 2 list of tables text][multi-shot]{
\begin{tabular}{|l|c|c|c|}
\hline
Methods  & Rank1 & Rank10 & Rank20 \\
\hline\hline
\textit{l}1-Graph\cite{UlGraph}  & 50.10&-&-\\
LOMO+XQDA\cite{LOMO} & 61.98 	&89.30		&93.62\\
CAMEL\cite{CAMEL}  &62.70 &-&-\\
MLAPG\cite{MLAPG}  & 64.24 &90.84&94.92\\
SSSVM\cite{SSSVM} &65.97&-&-\\
KNFST\cite{Zheng:nfst}&66.07&		91.56&		95.64\\
GOG+XQDA\cite{GOG}  & \color{blue}\textbf{67.28}		&\color{blue}\textbf{91.77}		&\color{blue}\textbf{95.93}\\
NK3ML(Ours) & \color{red}\textbf{76.77}	&\color{red}\textbf{95.58}&		\color{red}\textbf{98.02}\\
\hline \hline
**DGD\cite{DGD} & 66.60& -&-\\
*OLMANS\cite{OnlineNegSamples}  & 68.44&		92.67&	95.88\\
*SHaPE\cite{SHaPE} & 76.00 & -&-\\
*Spindle\cite{SpindleNet}  & \color{red}\textbf{79.90}&		\color{red}\textbf{97.10}&	\color{red}\textbf{98.60}\\
\hline
\end{tabular}
\label{table:CUHKM2}
}
\end{center}
\label{table:CUHK}
\end{table}

\setlength{\parskip}{1mm}
\noindent\textbf{Experiments on CUHK01 dataset:}
We use GOG and LOMO* as the features for CUHK01. Each person of the dataset has two images in each camera view. Hence we report comparison with both single-shot and multi-shot settings in Tables \ref{table:CUHKM1} and \ref{table:CUHKM2}. NK3ML provides the state-of-the-art performances in all ranks. For single-shot setting, it outperforms the current best method GOG+XQDA\cite{GOG} with a high margin of 9.20\%. Similarly for multi-shot setting, NK3ML improves the accuracy by  9.49\% for rank-1 over GOG+XQDA. At rank-1, NK3ML outperforms almost all of the pre/post processing based methods also, except DLPAR\cite{DLPAR} in single-shot setting, and Spindle\cite{SpindleNet} and SHaPE\cite{SHaPE} for multi-shot setting. However, note that Spindle and DLPAR uses other camera domain information for training, and SHaPE is a re-ranking technique to aggregate scores from multiple metric learning methods. Also note that NK3ML even outperforms the deep learning based methods
 (see Table \ref{table:viperall} also), 
emphasizing the limitation of deep learning based methods in re-ID systems with minimal training data. 
\setlength{\parskip}{0mm}

\setlength{\parskip}{1.5mm}
\noindent \textbf{Experiments on VIPeR dataset:}
Concatenated GOG, LOMO* and WHOS are used as the features for VIPeR. It is the most widely accepted benchmark for person re-ID. It is a very challenging dataset as it contains images captured from outdoor environment with large variations in background, illumination and viewpoint. An enormous number of algorithms have reported results on VIPeR, with most of them reporting an accuracy below 50\% at rank-1, as shown in Table \ref{table:viperall}. Even with the deep learning and pre/post processing re-ID methods, the best reported result for rank-1 is only 63.92\% by DCIA\cite{DCIA}. On the contrary, NK3ML provides unprecedented improvement over these methods and attains a 99.8\% rank-1 accuracy.
The superior performance of NK3ML is due to its capability to enhance the discriminability  even for the test data by simultaneously providing the maximal separation between the classes as well as minimizing the within class distance to the least value of zero. 
\setlength{\parskip}{0mm}

\begin{table}[h]
\begin{center}
\setlength{\abovecaptionskip}{9pt} 
\caption{Comparison with state-of-the-art results on VIPeR dataset. RN means Rank-N accuracy}
\label{table:viperall}
\resizebox{.99\columnwidth}{!}{%
\begin{tabular}{|l|c|c|c|c|}
\hline
Methods & Ref & R1 & R10 & R20 \\
\hline\hline
ELF\cite{ELF} & ECCV2008 & 12.0& 44.0& 61.0\\
PCCA\cite{PCCA} & CVPR2012 & 19.3 & 64.9 & 80.3\\
KISSME\cite{KISSME} &	CVPR2012   & 19.6 & 62.2  & 77.0 \\
LFDA\cite{LFDA:CVPR}	&	CVPR2013	&	24.2	&	67.1	&	-	\\
eSDC\cite{eSDC}	&	CVPR2013	&	26.7	&	62.4	&	76.4	\\
SalMatch\cite{SalMatch}	&	ICCV2013	&	30.2	&	-	&	-	\\
MLFL\cite{midlevel}	&	CVPR2014	&	29.1	&	66.0	&	79.9	\\
rPCCA\cite{rPcca} &	ECCV2014	& 22.0 & 71.0 & 85.3 \\
kLFDA\cite{rPcca} &	ECCV2014	& 32.3 & 79.7 & 90.9 \\
SCNCD\cite{colornames}	&	ECCV2014	&	37.8	&	81.2	&	90.4	\\
PolyMap\cite{ExPolyFeatMap}	&	CVPR2015	&	36.8	&	83.7	&	91.7	\\
LOMO+XQDA\cite{LOMO}	&	CVPR2015	&	40.0	&	80.5	&	91.1	\\
*Semantic\cite{Symantic}	&	CVPR2015	&	41.6	&	86.2	&	95.1	\\
QALF\cite{QALF}	&	CVPR2015	&	30.2	&	62.4	&	73.8	\\
CSL\cite{CSL}	&	ICCV2015	&	34.8	&	82.3	&	91.8	\\
MLAPG\cite{MLAPG}	&	ICCV2015	&	40.7	&	82.3	&	92.4	\\
*DCIA\cite{DCIA}	&	ICCV2015	&	\color{blue}\textbf{63.9}	&	87.5	&	-	\\
**DGD\cite{DGD}	&	CVPR2016	&	38.6	&	-	&	-	\\
KNFST\cite{Zheng:nfst}	&	CVPR2016	&	42.3 &	82.9	&	92.1 \\
\hline
\end{tabular}
\quad
    \begin{tabular}{|l|c|c|c|c|}
\hline
Methods & Ref & R1 & R10 & R20 \\
\hline\hline
SSSVM\cite{SSSVM}	&	CVPR2016	&	42.1	&	84.3	&	91.9	\\
**TPC\cite{TPC}	&	CVPR2016	&	47.8	&	84.8	&	91.1	\\
GOG+XQDA\cite{GOG}	&	CVPR2016	&	49.7	&	88.7	&	94.5	\\
SCSP\cite{SCSP}	&	CVPR2016	&	53.5	&	\color{blue}\textbf{91.5}	&	\color{blue}\textbf{96.7}	\\
**SCNN\cite{SCNN}	&	ECCV2016	&	37.8	&	66.9	&	-	\\
**Shi et al.\cite{Shi}	&	ECCV2016	&	40.9	&	-	&	-	\\
\textit{l}1-graph\cite{UlGraph}	&	ECCV2016	&	41.5	&	-	&	-	\\
**S-LSTM\cite{SLSTM}	&	ECCV2016	&	42.4	&	79.4	&	-	\\
*SSDAL\cite{SSDAL}	&	ECCV2016	&	43.5	&	81.5	&	89.0\\
*TMA\cite{TMA}	&	ECCV2016	&	48.2	&	87.7	&	93.5	\\
*SSM\cite{song:scalableManifold}	&	CVPR2017	&	53.7	&	\color{blue}\textbf{91.5}	&	96.1   \\
*Spindle\cite{SpindleNet} & CVPR2017 & 53.8	&	83.2	&	92.1\\
CAMEL\cite{CAMEL} & ICCV2017 & 30.9 &	-	&	-	\\
*MuDeep & ICCV2017 & 43.0	&	85.8	& - \\
*OLMANS\cite{OnlineNegSamples} & ICCV2017 & 45.0	&	85.0	&	93.6\\
*DLPAR\cite{DLPAR} & ICCV2017 & 48.7	&	85.1	&	93.0\\
*PDC\cite{PDC} & ICCV2017 & 51.3	&	84.2	&	91.5\\
*SHAPE\cite{SHaPE} & ICCV2017 & 62.0 &	-	&	-	\\
\hline\hline
NK3ML &  Ours & \color{red}\textbf{99.8}	&	\color{red}\textbf{100}	& 	\color{red}\textbf{100}\\
\hline
\end{tabular}
}
\end{center}
\end{table}

\begin{table}[h]
\begin{center}
\setlength{\abovecaptionskip}{9pt} 
\caption{Comparison of execution time (in seconds) on VIPeR dataset}
\begin{tabular}{|l|c|c|c|c|c|c|c|c|}
\hline
Methods & NK3ML & NFST& KNFST & XQDA & MLAPG   &  
kLFDA   & MFA  &  rPCCA\\ 
\hline\hline
Training & 	1.64 & 1.47	& 0.37	 & 1.35	&	12.10 &	4.10&	3.68&23.98\\
Testing & 0.37	& 0.34 	& 0.33	&	0.34 & 0.13	&	4.13&	3.99& 3.74\\
\hline
\end{tabular}
\label{table:executionTime}
\end{center}
\end{table}

\subsection{Computational Requirements}

We compare the execution time of 
NK3ML with other metric learning methods including NFST\cite{Zheng:nfst}, KNFST\cite{Zheng:nfst}, XQDA\cite{LOMO,GOG},  MLAPG\cite{MLAPG}, kLFDA\cite{rPcca}, MFA\cite{rPcca} and rPCCA\cite{rPcca}  on VIPeR dataset. The details are shown in Table \ref{table:executionTime}. The training time is calculated for the 632 samples in the training set, and the testing time is calculated for all the 316 queries in the test set. The training and testing time are averaged over 10 random trials. All methods are implemented in MATLAB
on a PC with an Intel i7-6700 CPU@3.40GHz and 32GB memory. The testing time for NK3ML is 0.37s for the set of 316 query images (0.0012s per query), which is adequate for real time applications.

\begin{figure}[h] 
\begin{center}
  \subfloat[]{%
    \includegraphics[width=0.35\linewidth]{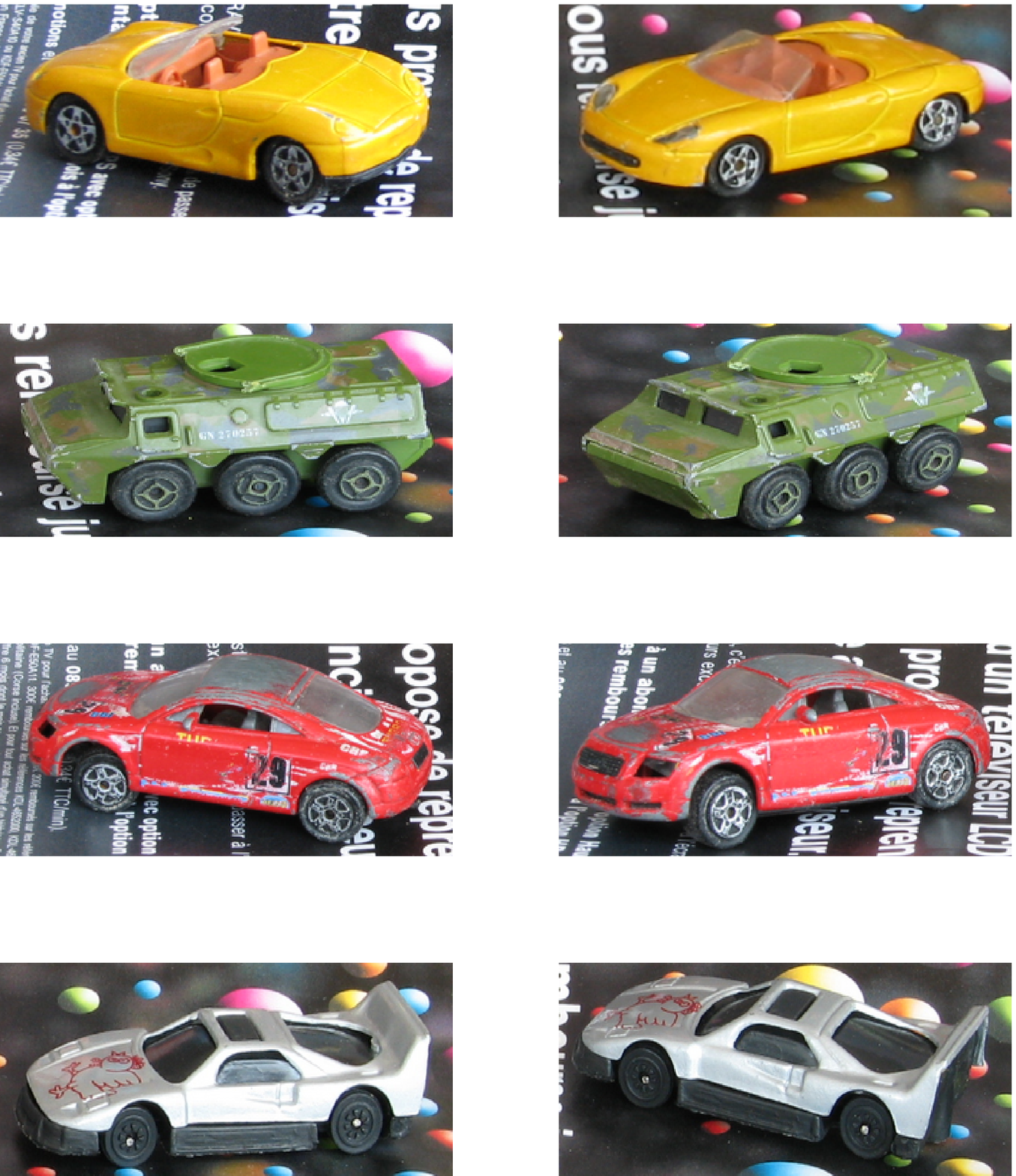}
  } 
 \qquad
  \subfloat[]{%

   \includegraphics[width=0.5\linewidth]{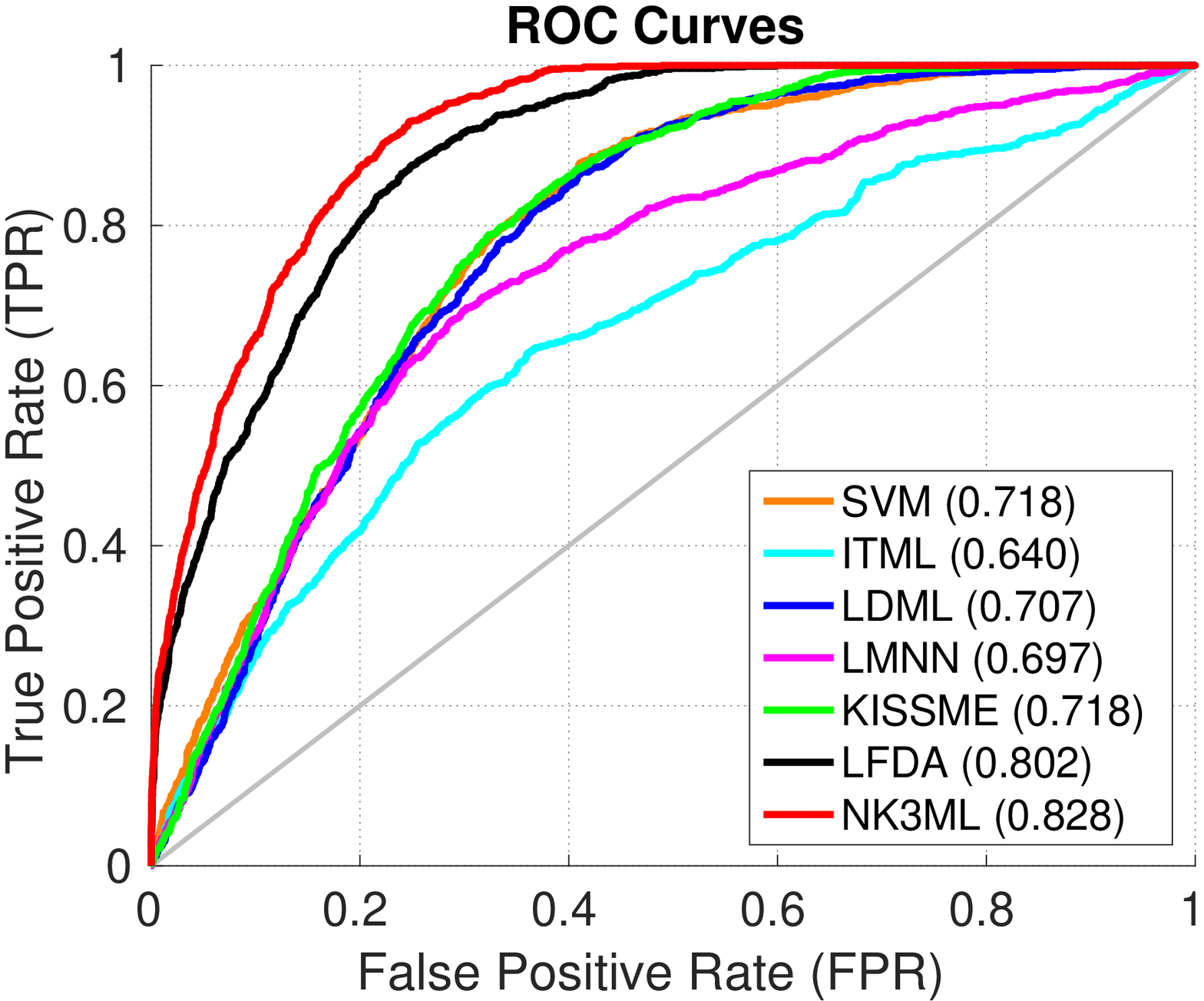}
  } 
  \end{center}
  \caption{ToyCars dataset (a) Sample images (b) ROC curves and EER comparisons.} 
  \label{fig:ToyCar}
\end{figure}

\subsection{ Application in Another Domain}
In order to evaluate the applicability of NK3ML on other object verification problems also, we conduct experiments using LEAR ToyCars \cite{ToyCarDataset} dataset. It contains a total of 256 images of 14 distinct cars and trucks. The images have wide variations in pose, illumination and background. The objective is to verify if a given pair of images are similar or not, even if they are \textit{unseen} before. The training set has 7 distinct objects, provided as 1185 similar pairs and 7330 dissimilar pairs. The remaining 7 objects are used in the test set with 1044 similar pairs and 6337 dissimilar pairs. We use  the feature representation from \cite{KISSME}, which uses LBP with HSV and Lab histograms. 

We compare the performance of NK3ML with the state-of-the-art metric learning methods including KISSME\cite{KISSME}, ITML\cite{ITML}, LDML\cite{LDML}, LMNN\cite{LMNN1,LMNN2}, LFDA\cite{LFDA:ICML,LFDA:CVPR} and SVM\cite{SVM}. Note that NK3ML and LMNN need the true class labels (not the similar/dissimilar pairs) for training. The proposed NK3ML learned a six dimensional subspace. For fair comparisons, we use the same features and learn an equal dimensional subspace for all the methods. We plot the Receiver Operator Characteristic (ROC) curves of the methods in Fig. \ref{fig:ToyCar}, with the Equal Error Rate (EER) shown in parenthesis. NK3ML outperforms all other methods with a good margin. This experiment re-emphasizes that NK3ML is efficient to generalize well for unseen objects. Moreover, it indicates that NK3ML has the potential for other object verification problems also, apart from person re-identification.

\color{black}
\section{Conclusions}
\label{sec:Conclusion}
In this work we presented a novel metric learning framework to efficiently address the small training sample size problem inherent in re-ID systems due to high dimensional data.  We identify the suboptimality of NFST in generalizing to the test data. We provide a solution that  minimizes the intra-class distance of training samples trivially to zero, as well as maximizes the inter-class distance to a much higher margin so that the learned discriminant vectors are effective in terms of generalization of the classifier performance for the test data also. Experiments on various challenging benchmark datasets show that our method outperforms the state-of-the-art metric learning approaches. Especially, our method attains near human level perfection in the most widely accepted dataset VIPeR. We evaluate our method on another object verification problem also and validate its efficiency to generalize well to unseen data.\\

\noindent \textbf{Acknowledgement.} This research work is supported by Ministry of Electronics and Information Technology (MeitY), Government of India, under  Visvesvaraya PhD Scheme. 

\bibliographystyle{splncs04}
\bibliography{TMFerozAli}
\end{document}